\pdfoutput=1

\documentclass[11pt]{article}

\usepackage[preprint]{acl}

\usepackage{times}
\usepackage{latexsym}

\usepackage[T1]{fontenc}

\usepackage[utf8]{inputenc}

\usepackage{microtype}

\usepackage{inconsolata}

\usepackage{graphicx}

\usepackage{algorithm}
\usepackage{algpseudocode}
\usepackage{amsmath}
\usepackage{hyperref}

\title{A Multi-AI Agent System for Autonomous Optimization of Agentic AI Solutions via Iterative Refinement and LLM-Driven Feedback Loops}

\author{Kamer Ali Yuksel \and Hassan Sawaf \\
        aiXplain Inc., San Jose, CA, USA \\
  \texttt{\{kamer, hassan\}@aixplain.com} \\}

\begin{document}
\maketitle
\begin{abstract}
Agentic AI systems use specialized agents to handle tasks within complex workflows, enabling automation and efficiency. However, optimizing these systems often requires labor-intensive, manual adjustments to refine roles, tasks, and interactions. This paper introduces a framework for autonomously optimizing Agentic AI solutions across industries, such as NLP-driven enterprise applications. The system employs agents for Refinement, Execution, Evaluation, Modification, and Documentation, leveraging iterative feedback loops powered by an LLM (\emph{Llama 3.2-3B}). The framework achieves optimal performance without human input by autonomously generating and testing hypotheses to improve system configurations. This approach enhances scalability and adaptability, offering a robust solution for real-world applications in dynamic environments. Case studies across diverse domains illustrate the transformative impact of this framework, showcasing significant improvements in output quality, relevance, and actionability. All data for these case studies, including original and evolved agent codes, along with their outputs, are here:  \href{https://anonymous.4open.science/r/evolver-1D11/}{anonymous.4open.science/r/evolver-1D11/} 
\end{abstract}

\section{Introduction}
Agentic AI systems, composed of specialized agents working collaboratively to achieve complex objectives, have transformed industries such as market research, business process optimization, and product recommendation. These systems excel in automating decision-making and streamlining workflows. However, their optimization remains challenging due to the complexity of agent interactions and reliance on manual configurations.

Recent advancements in large language models (LLMs) provide a solution by enabling automated refinement of Agentic AI systems. LLMs can autonomously generate and evaluate complex hypotheses, facilitating iterative improvements in agent roles and workflows with minimal human oversight. Case studies conducted demonstrate how these advancements address domain-specific challenges. These examples highlight the framework's scalability and adaptability, making it particularly effective for dynamic, evolving environments.

This paper presents a framework for autonomous optimization of Agentic AI systems using LLM-powered feedback loops. The framework improves efficiency and scalability by refining agent configurations based on qualitative and quantitative metrics. Several case studies across various domains provide evidence of the framework's ability to overcome domain-specific challenges. Designed for deployment in enterprise systems, this framework addresses persistent challenges in optimizing complex workflows in real-world settings.

This work establishes a scalable, autonomous system for optimizing Agentic AI with broad applicability across industries. \textbf{Key contributions:}
\begin{itemize} \item \textbf{Evolutionary Optimization:} Evolving agent configurations without manual intervention. \item \textbf{Autonomous Refinement:} Fully automated optimization through iterative feedback loops. \item \textbf{Validation Case Studies:} Empirical results from case studies in diverse domains, demonstrating significant performance gains. \end{itemize}

\section{Background}
Agentic AI systems automate complex processes across industries, providing significant efficiency gains. However, their optimization requires addressing the intricacies of agent interactions, particularly in dynamic environments with evolving objectives. Recent advancements in LLMs offer transformative capabilities by enabling autonomous generation and evaluation of hypotheses to improve workflows. This framework distinguishes itself by enabling fully autonomous optimization of Agentic AI systems. The system enhances scalability, adaptability, and domain independence through LLM-driven feedback loops, hypothesis generation, and iterative modifications, setting a new standard for optimizing complex AI workflows. Unlike previous approaches reliant on predefined tasks or manual intervention, this method offers a superior solution for real-world, dynamic environments.

Prior research has explored various aspects of agentic systems' optimization and LLM integration. For instance, \citet{liu2024mlagentbench} introduced \textit{MLAgentBench}, a benchmark for evaluating language agents across diverse tasks. While this framework provides valuable insights, it focuses primarily on performance evaluation rather than iterative workflow refinement, which is our study's focus. Similarly, \citet{smith2023largemodelagents} explored the use of \textit{Large Model Agents (LMAs)} to enhance cooperation between agents using LLMs, highlighting the potential of LLMs for iterative feedback loops among agents. This aligns closely with the current study’s emphasis on autonomous refinement processes. 

\citet{johnson2023professionalagents} demonstrated how LLMs enable agents to autonomously refine their roles and workflows, underscoring the significance of optimizing agentic AI systems. Meanwhile, \citet{pan2024autonomousevaluation} proposed using automated evaluators to refine agent performance in tasks like web navigation. While effective in that domain, it lacks the scalability and domain independence targeted by the current framework. Similarly, \citet{skilldiscovery2024agentic} introduced an LLM-based skill discovery framework, resonating with the iterative task proposals discussed in this study. Furthermore, \citet{hu2024automated} emphasized the importance of modular components and foundational models for planning and executing agentic systems, focusing on the design phase, whereas the current study emphasizes continuous refinement and workflow optimization.

\citet{masterman2024landscape} reviewed emerging AI agent architectures, focusing on modularity and scalability, which are central to this framework’s refinement approach. \citet{agentinstruct2024} proposed a framework for generating synthetic data with agents, showcasing their potential to refine outputs via feedback loops. \citet{improvingagents2024reflectivetree} introduced the \textit{Reflective Tree} for multistep decision-making, aligning with this study’s iterative design. \citet{feedbackhacking2024} addressed feedback loop risks like reward hacking, which this framework mitigates through robust evaluation. Finally, \citet{aiagents2024} emphasized agent benchmarks, reinforcing this framework’s focus on qualitative metrics.

Recently, Automated Design of Agentic Systems (ADAS) was introduced by \citet{hu2024automated}, focusing on creating new agentic designs through a meta-agent that programs novel agents by combining and refining building blocks. While ADAS is designed to invent new agents, the framework presented in this study is focused on optimizing existing agent systems. Through iterative LLM-driven feedback loops, agent roles, tasks, and workflows are refined to enhance adaptability and scalability in dynamic environments. Unlike ADAS, which prioritizes agent creation, this work focuses on continuously improving and optimizing established systems.

\begin{algorithm}
\caption{Agentic AI Refinement Process}
\begin{algorithmic}[1]

\State \textbf{Input:}
\State \hspace{2em} $C_0$: Initial code
\State \hspace{2em} $\text{criteria}$: Qualitative evaluation criteria
\State \hspace{2em} $\epsilon$: Improvement threshold
\State \hspace{2em} $\text{max\_iterations}$: Maximum number of iterations

\State \textbf{Output:}
\State \hspace{2em} $C_{\text{best}}$: Best-known code variant
\State \hspace{2em} $O_{\text{best}}$: Output of the best-known code variant

\State \textbf{Initialization:}
\State $C_{\text{best}} \gets C_0$
\State $O_{\text{best}} \gets \text{execute}(C_0)$
\State $S_{\text{best}} \gets f(O_{\text{best}}, \text{criteria})$
\State $\text{iteration} \gets 0$

\While{$\text{iteration} < \text{max\_iterations}$}
    \State $\text{iteration} \gets \text{iteration} + 1$
    
    \State $E_{\text{best}} \gets \text{evaluate}(O_{\text{best}}, \text{criteria})$
    \State $\mathcal{H}_i \gets \text{generate\_hypotheses}(E_{\text{best}})$
    \State $C_{i+1} \gets M(\mathcal{H}_i, C_{\text{best}})$
    
    \State $O_{i+1} \gets \text{execute}(C_{i+1})$
    \State $S_{i+1} \gets f(O_{i+1}, \text{criteria})$
    
    \If{$S_{i+1} > S_{\text{best}}$}
        \State $C_{\text{best}} \gets C_{i+1}$
        \State $O_{\text{best}} \gets O_{i+1}$
        \State $S_{\text{best}} \gets S_{i+1}$
        \State \text{Save best-known variant and output}
    \EndIf
    
    \If{$|S_{i+1} - S_{\text{best}}| < \epsilon$}
        \State \textbf{break} \Comment{Stop if improvement is below threshold}
    \EndIf
\EndWhile

\State \textbf{Return} $C_{\text{best}}, O_{\text{best}}$

\end{algorithmic}
\end{algorithm}

\newpage

\section{Architecture}
The proposed method for autonomous refinement and optimization of Agentic AI systems leverages several specialized agents, each responsible for a specific phase in the refinement process. This method operates in iterative cycles, continuously refining agent roles, goals, tasks, workflows, and dependencies based on qualitative and quantitative output evaluation. Moreover, the system is designed with scalability, ensuring its deployment across industries. The LLM-driven feedback loops offer a foundational infrastructure for adapting the system to various NLP applications, ensuring broad applicability across domains. The proposed method’s optimization process is guided by two core frameworks: the Synthesis Framework and the Evaluation Framework. Synthesis Framework framework generates hypotheses based on the system’s output. The Hypothesis Agent and Modification Agent collaborate to synthesize new configurations for the Agentic AI system, proposing modifications to agent roles, goals, and tasks; to be tested by the Evaluation Framework.

The refinement and optimization process is structured into these frameworks, contributing to the continuous improvement of the Agentic AI solution. The proposed method operates autonomously, iterating through cycles of hypothesis generation, execution, evaluation, and modification until optimal performance is achieved. A detailed report of a refinement iteration is provided in Appendix \ref{sec:appendix}. This method begins by deploying a baseline version of the Agentic AI system. Agents are assigned predefined roles, tasks, and workflows, and the system generates initial qualitative and quantitative criteria based on the system’s objectives. An LLM is used to analyze the system’s code and extract evaluation metrics, which serve as benchmarks for assessing future outputs. Human input can be introduced to revise or fine-tune the evaluation criteria to better align with project goals; which is optional, as the method is designed to operate autonomously.

The proposed method begins with a baseline version of the Agentic AI system, assigning initial agent roles, goals, and workflows. The first execution is run to generate the initial output and establish the baseline for comparison. After evaluating the initial output, the Hypothesis Agent generates hypotheses for modifying agent roles, tasks, or workflows based on the evaluation feedback. These hypotheses are then passed to the Modification Agent, which synthesizes changes to agent logic, interactions, or dependencies, producing new system variants. The Execution Agent executes the newly modified versions of the system, and performance metrics are gathered. The outputs generated are evaluated using qualitative and quantitative criteria (e.g., clarity, relevance, execution time). The Selection Agent compares the newly generated outputs against the best-known variant, ranks the variants, and determines whether the new output is superior. Memory Module stores The best-performing variants for future iterations. The cycle repeats as the proposed method continues refining the agentic workflows, improving overall performance until predefined or generated (and optionally revised) criteria are satisfied.

\subsection{Synthesis Framework}
The Refinement Agent manages the iterative optimization process by delegating tasks to other agents and synthesizing hypotheses for improving the system. It evaluates agent outputs against qualitative and quantitative criteria, identifying areas where agent roles, tasks, or workflows can be improved. The Refinement Agent leverages evaluation metrics such as clarity, relevance, depth of analysis, and actionability to propose modifications that enhance system output. The Hypothesis Generation Agent proposes specific changes to the agent system based on the output analysis. This module generates hypotheses for improving agent roles, tasks, and interactions based on evaluation feedback. For example, if agents are underperforming due to inefficiencies in their task delegation, the hypothesis module might suggest altering task hierarchies or reassigning specific roles.

The Modification Agent implements changes based on the hypotheses generated by the Refinement Agent. These changes may involve adjusting agent logic, modifying workflows, or altering agent dependencies. By synthesizing these changes, our method creates multiple variants of the Agentic AI solution. Each variant is stored and documented, with details regarding the expected improvements. The Execution Agent runs modified versions of the system, executing the newly generated variants and collecting performance data for subsequent evaluation. It ensures that agents perform their tasks as specified in the new configuration and debug issues as they arise. The Execution Agent tracks qualitative and quantitative outputs, feeding this information into the evaluation process.

\subsection{Evaluation Framework}
The Evaluation Framework is responsible for assessing the outputs of each system variant. The Evaluation Agent employs Llama 3.2-3B to evaluate both qualitative and quantitative aspects of the system’s performance. The Evaluation Framework ensures that each iteration aligns with the system’s overarching objectives, focusing on continuous improvement. The Evaluation Agent assesses the outputs of each system variant using a LLM. The LLM evaluates outputs based on predefined or generated qualitative criteria, including clarity, relevance to the task, depth of analysis, actionability, and quantitative metrics such as execution time and success rate. The Evaluation Agent provides a comprehensive system performance analysis, identifying areas for further improvement. After each iteration, the Selection Agent compares the outputs of the modified system against the best-known configuration. It ranks the new variants based on the evaluation scores provided by the Evaluation Agent, determining which configuration yields the highest performance. The top-ranked variant is stored for future iterations, ensuring continuous improvement.

\subsection{Refinement Process}
Agentic AI refinement process begins with the initialization of the best-known code variant, denoted as \( C_0 \), and the generation of its corresponding output, \( O_{C_0} \). The performance of the output is evaluated using a set of qualitative criteria (e.g., clarity, relevance, depth of analysis), where the evaluation function \( f(O_C, \text{criteria}) \) produces a score \( S(C_0) = f(O_{C_0}, \text{criteria}) \) based on these criteria. This initial score, \( S(C_0) \), is the baseline for comparison in subsequent iterations. At each iteration \( i \), the current best-known output, \( O_{C_i} \), is evaluated, and a set of hypotheses, \( \mathcal{H}_i = \text{generate\_hypotheses}(E_{C_i}) \), is generated from the qualitative evaluation \( E_{C_i} \) to suggest improvements. The hypotheses \( \mathcal{H}_i \) are then applied to the code \( C_i \), resulting in a new variant \( C_{i+1} = M(\mathcal{H}_i, C_i) \). The new code variant \( C_{i+1} \) is executed, producing a new output \( O_{C_{i+1}} \). The new output is evaluated using the same evaluation function \( f(O_C, \text{criteria}) \), yielding a new score \( S_{i+1} = f(O_{C_{i+1}}, \text{criteria}) \). If the new score \( S_{i+1} \) is greater than the best-known score \( S_{\text{best}} = \max(S_{i+1}, S_{\text{best}}) \), the new variant is considered superior, and the best-known variant is updated as follows. The process continues iteratively until a stopping condition is met, either when the improvement between iterations becomes smaller than a predefined threshold \( |S_{i+1} - S_{\text{best}}| < \epsilon \), or when a maximum number of iterations is reached. Upon termination, the proposed method returns the best-known variant \( C_{\text{best}} \) and its output \( O_{\text{best}} \).

Once initialized, the proposed method enters the execution phase, where agents perform their assigned tasks according to the baseline configuration. The Execution Agent runs the system, producing initial outputs that serve as a baseline for comparison in subsequent iterations. The results of this execution phase are stored for future analysis and comparison. The Evaluation Agent evaluates the outputs produced in the execution phase. The proposed method employs qualitative and quantitative criteria to assess the quality of the outputs. Qualitative metrics include relevance, clarity, depth of analysis, and actionability, while quantitative metrics include execution time, task completion rate, and overall system efficiency. The Evaluation Agent uses an LLM to generate detailed feedback, identifying areas where the system can be improved. The Hypothesis Generation Agent analyzes the evaluation data, generating hypotheses for improving agent roles, tasks, and workflows. These hypotheses may include changes such as altering task delegation, modifying agent goals, or restructuring the interdependencies between agents. Once the hypotheses are generated, the Modification Agent implements the proposed changes, creating new system variants based on these modifications. The modified versions of the system are re-executed by the Execution Agent, and the Evaluation Agent again evaluates their outputs. This iterative process continues, with each new variant compared against the previous best-known configuration. The Selection Agent ranks the system variants based on performance, ensuring that the Memory Module only stores the top-performing versions.

\section{Case Studies}

The evolution of agent systems in various domains highlights the need for continuous refinement to meet the dynamic demands of industry standards and user expectations. This section presents an overview of several case studies that illustrate the transformative process of refining agent systems across diverse applications, including market research, AI architecting, career transitions, outreach strategies, LinkedIn posts, meeting facilitation, lead generation, content creation, and presentation development. Each case study showcases the challenges faced by the original systems, the strategic modifications implemented, and the resultant improvements in output quality. The findings underscore the significance of specialization and data-driven decision-making in enhancing agent system performance. All data for these case studies, including original and evolved agent codes, along with their outputs and evaluation reports, are here:  \href{https://anonymous.4open.science/r/evolver-1D11/}{anonymous.4open.science/r/evolver-1D11/} 

\subsection{Market Research Agent}

The original market research agent system was developed to provide strategic insights. However, it encountered several challenges, including inadequate market research depth, subpar strategy development, and limited output quality. These deficiencies hindered the system's ability to effectively align with user needs, resulting in low scores across evaluation criteria. The evolved agent system introduced specialized roles such as Market Research Analyst, Data Analyst, and User Experience Specialist to address these issues. These changes aimed to enhance the depth of market analysis, create a data-driven decision-making framework and prioritize user-centered design principles. The system was better equipped to understand emerging trends and deliver actionable insights by incorporating specialized agents. The refined agent system achieved remarkable improvements in output quality, scoring 0.9 in alignment and relevance, accuracy and completeness, and clarity and actionability. The evolved outputs provided a coherent strategy framework, significantly enhancing the overall effectiveness of the market research agent.

\begin{figure}[h]
\centering
\includegraphics[width=0.5\textwidth]{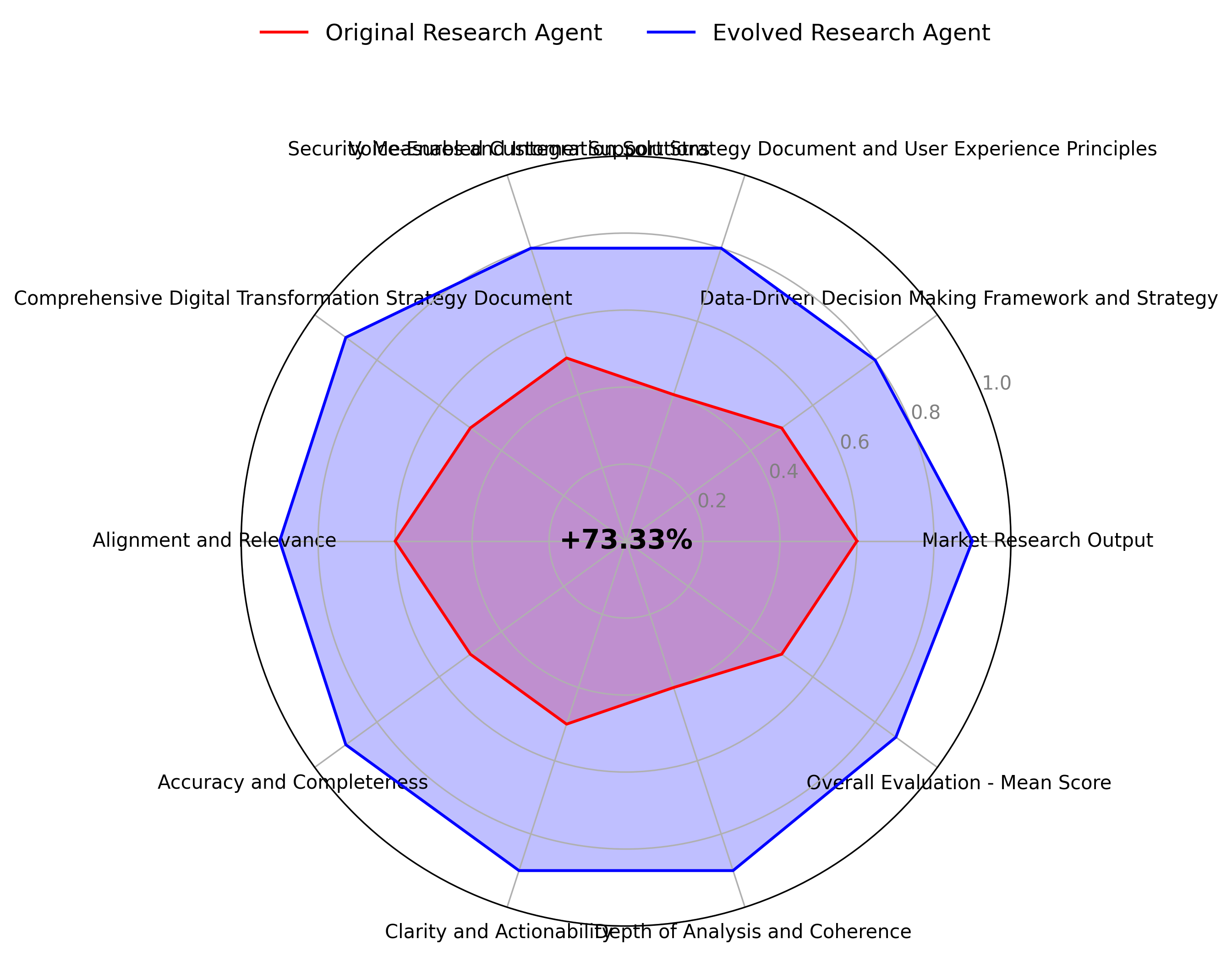}
\caption{Market Research Agent Refinement}
\end{figure}

\subsection{Medical AI Architect Agent}

The architect agent system for medical imaging faced challenges related to regulatory compliance, patient engagement, and explainability of AI-driven decision-making processes. These limitations resulted in moderate evaluation scores, undermining the system's effectiveness in addressing critical healthcare needs. In response, the evolved system incorporated specialized agents, including a Regulatory Compliance Specialist and a Patient Advocate, to ensure adherence to standards and prioritize patient needs. Developing transparency frameworks strengthened the focus on explainability, while continuous monitoring mechanisms were established for ongoing performance assessment. The evolved system's outputs demonstrated significant improvements across multiple evaluation criteria, including regulatory compliance (0.9), patient-centered design (0.8), and explainability (0.8). These enhancements underscore the importance of specialization in developing systems that meet complex healthcare demands, ultimately leading to improved patient care and outcomes.

\begin{figure}[h]
\centering
\includegraphics[width=0.5\textwidth]{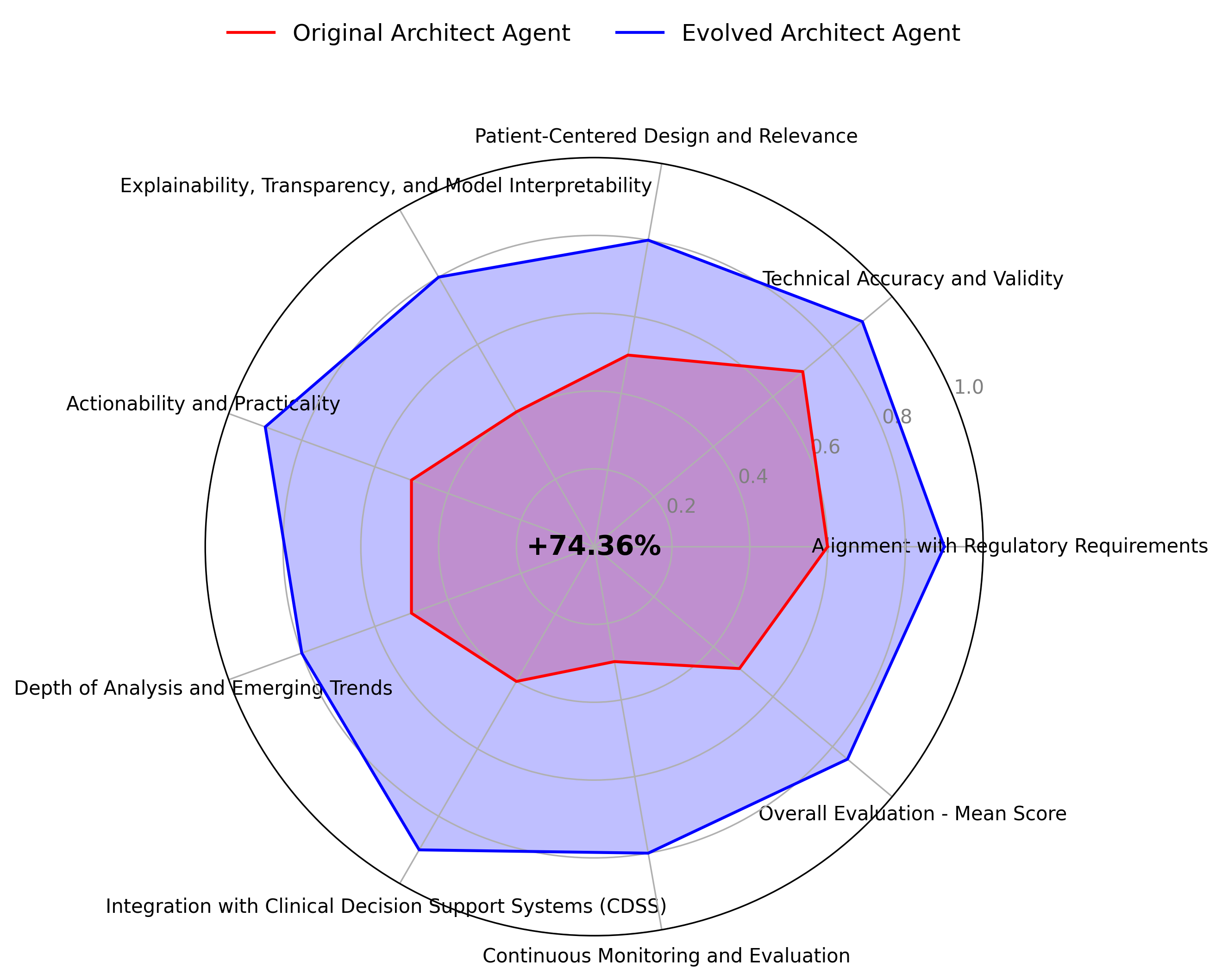}
\caption{AI Architect Agent Refinement}
\end{figure}

\subsection{Career Transition Agent}

The original AI transition agent system was intended to assist software engineers in transitioning to AI specialist roles. However, it struggled with alignment to industry expertise and clarity in career growth goals. This disconnect resulted in ineffective action plans and poor communication clarity. The evolved system adopted a multifaceted approach, introducing new agents such as Domain Specialist and Skill Developer. The tasks were refined to ensure specificity and clarity, enhancing communication through detailed timelines and structured outputs. The modifications led to substantial improvements in evaluation scores, with notable advancements in alignment with AI domain expertise (91\%) and communication clarity (90\%). The enhanced system provides clear, actionable goals, facilitating a more effective transition for software engineers into AI roles and highlighting the importance of agent system refinement.

\begin{figure}[h]
\centering
\includegraphics[width=0.5\textwidth]{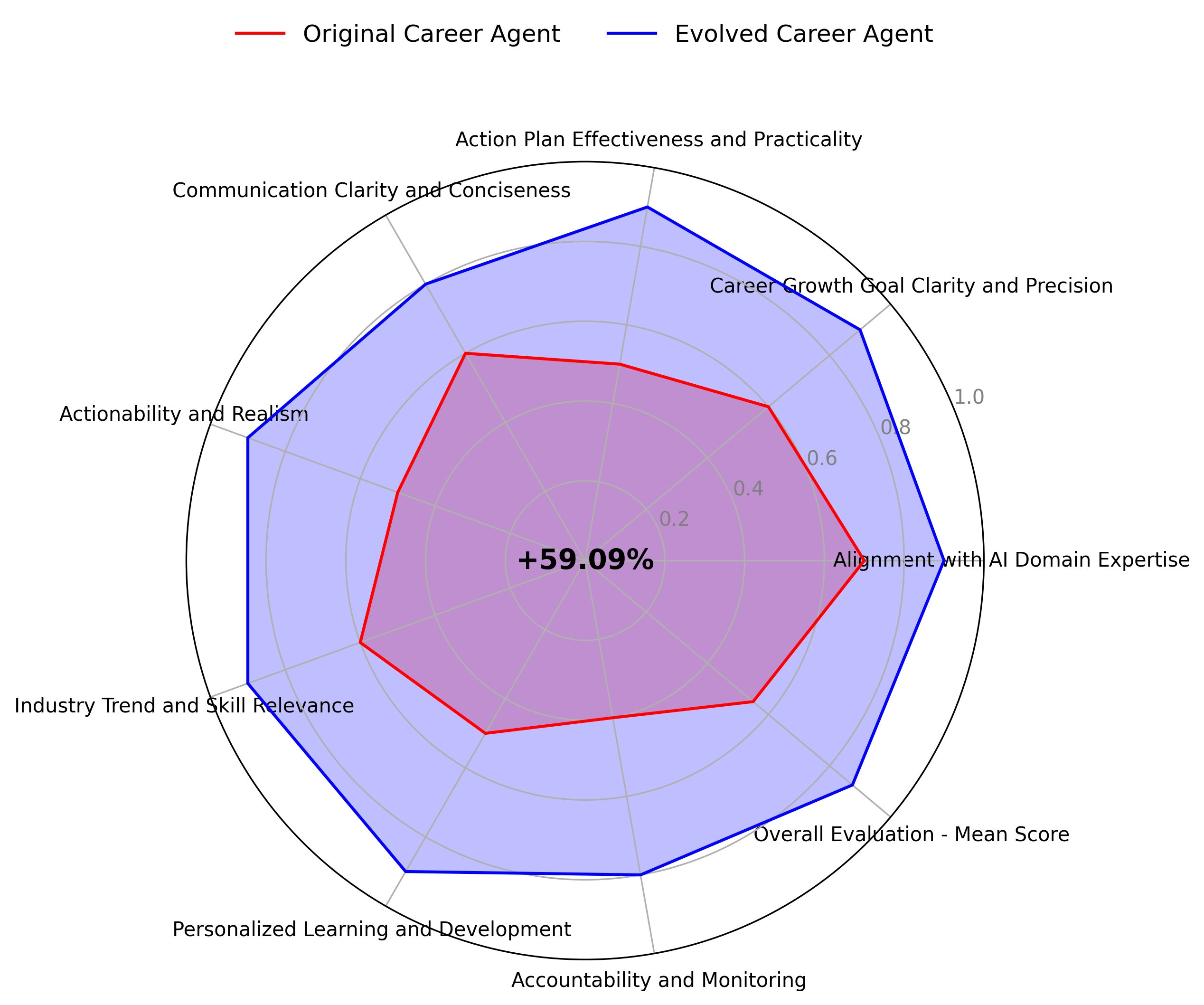}
\caption{Career Transition Agent Refinement}
\end{figure}

\subsection{Outreach Agent}

Initially, the outreach agent system designed for the supply chain faced limitations due to its narrow focus and poor output quality. The original system was characterized by basic roles, such as Email Drafter, which failed to address the complexities of supply chain management. Five specialized roles were introduced to enhance the system, focusing on supply chain analysis, optimization, and sustainability. This comprehensive approach allowed for a deeper analysis of supply chain challenges and operational inefficiencies. The evolved agent system demonstrated significant improvements, with enhanced clarity, accuracy, and actionability in outputs. The modifications led to outputs that exceeded the refined evaluation criteria, establishing the system as a valuable tool for e-commerce companies seeking effective supply chain solutions.

\begin{figure}[h]
\centering
\includegraphics[width=0.5\textwidth]{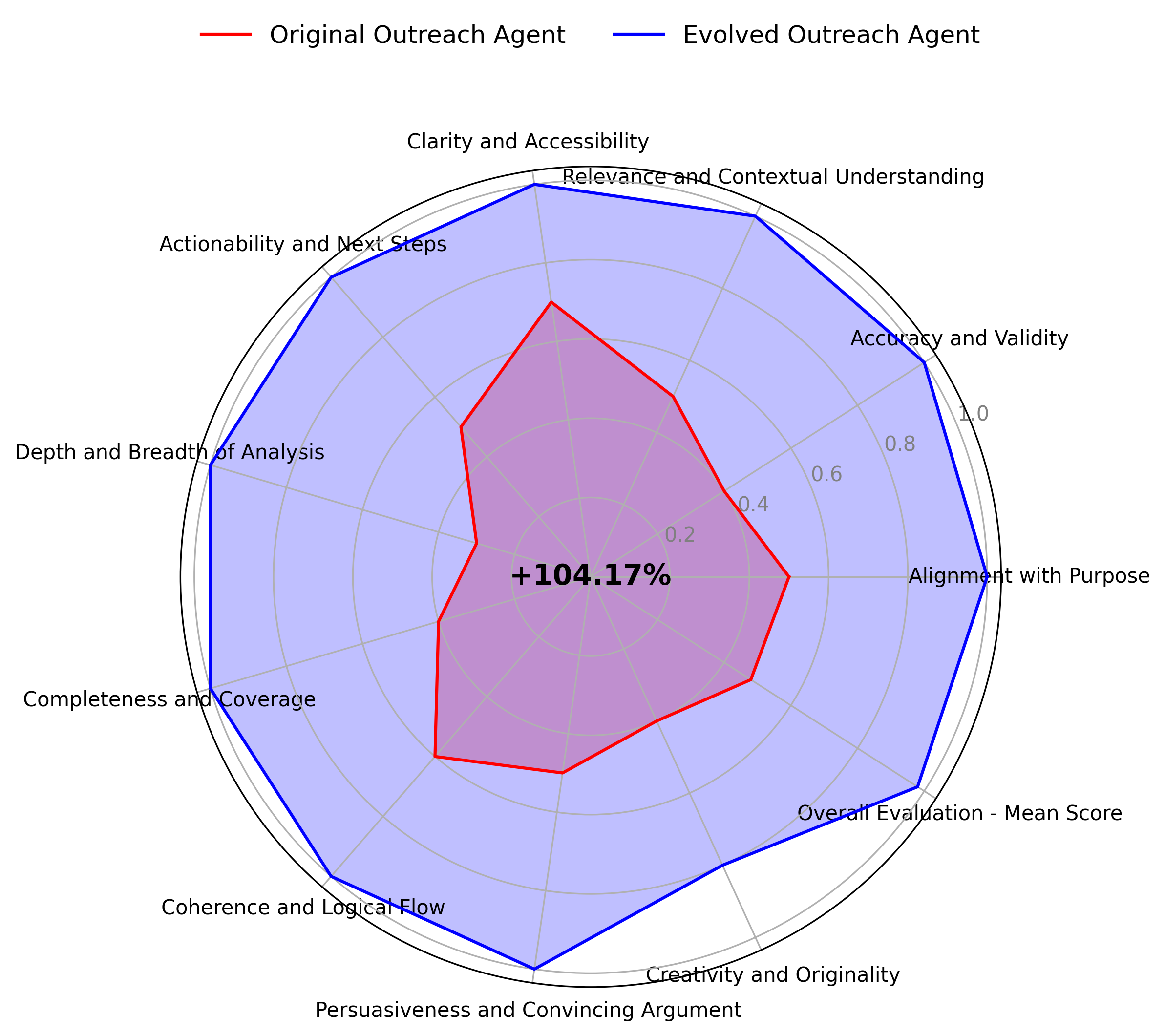}
\caption{Outreach Agent Refinement}
\end{figure}

\subsection{LinkedIn Agent}

The original generative AI agent system struggled with limitations in-depth, audience engagement, and source credibility when creating LinkedIn posts on generative AI trends. These challenges affected the system's ability to generate insightful and engaging content. The evolved system incorporated four specialized roles, including an Audience Engagement Specialist, to enhance content development and audience interaction. To ensure relevancy, a dynamic content strategy emphasizing audience metrics and adaptability was implemented. The refined outputs significantly improved contextual relevance, accuracy, audience engagement potential, and clarity. The enhanced system positioned itself as a valuable resource for stakeholders interested in generative AI trends, highlighting the importance of specialized roles in content creation.

\begin{figure}[h]
\centering
\includegraphics[width=0.5\textwidth]{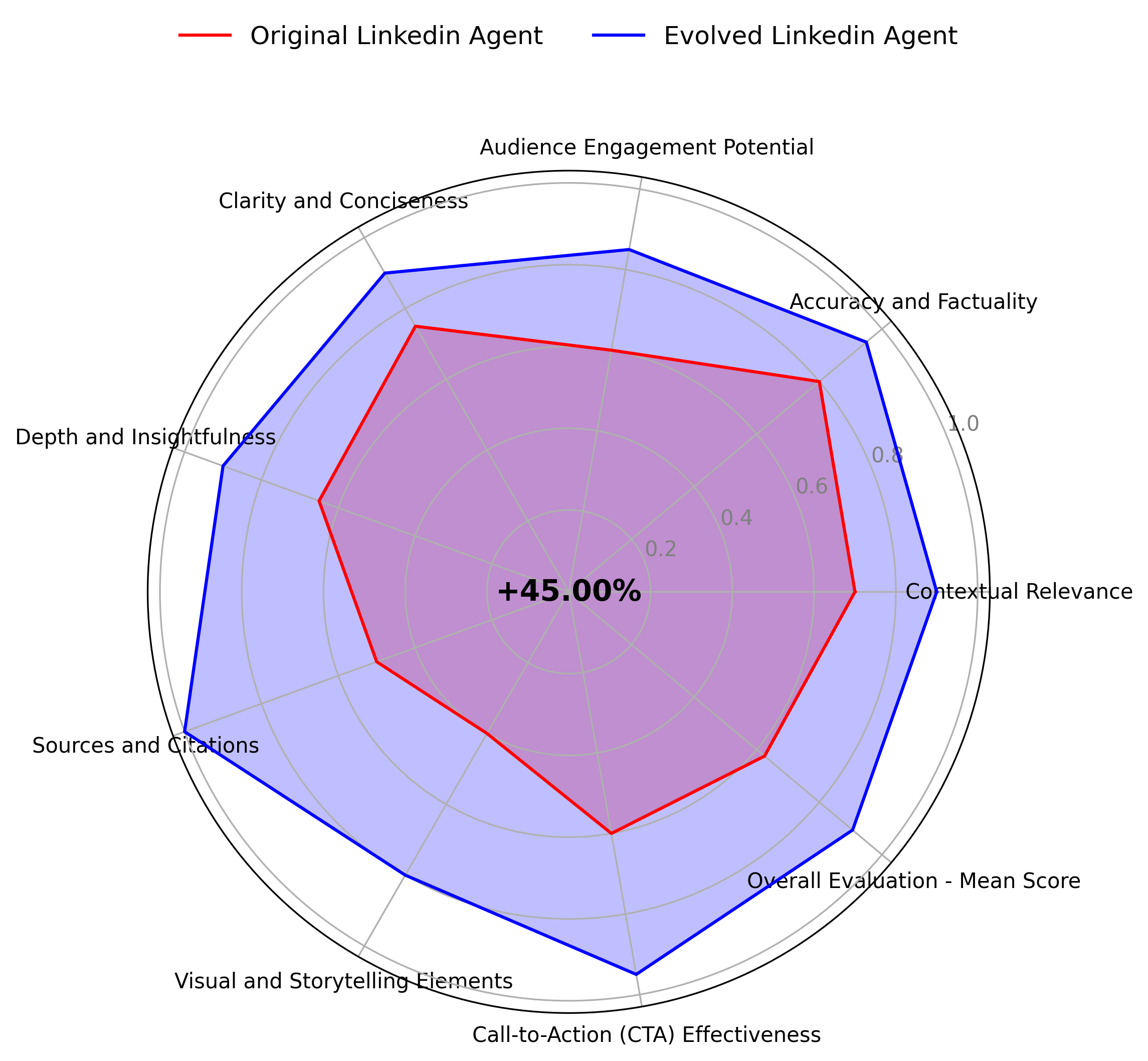}
\caption{LinkedIn Agent Refinement}
\end{figure}

\subsection{Meeting Agent}

The meeting agent system designed for AI-powered drug discovery did not meet qualitative evaluation criteria due to poor alignment with industry trends and insufficient analytical depth. These shortcomings limited its effectiveness in supporting pharmaceutical stakeholders. The evolved system introduced specialized roles, including AI industry experts and regulatory compliance leads, to provide comprehensive insights and ensure outputs were aligned with stakeholder needs. This overhaul aimed to enhance the system's relevance and actionability. The comparison of outputs revealed substantial improvements, with the evolved system achieving scores of 0.9 or higher across all evaluation categories. The refined system effectively addressed the needs of the pharmaceutical industry, demonstrating the impact of targeted modifications.

\begin{figure}[h]
\centering
\includegraphics[width=0.5\textwidth]{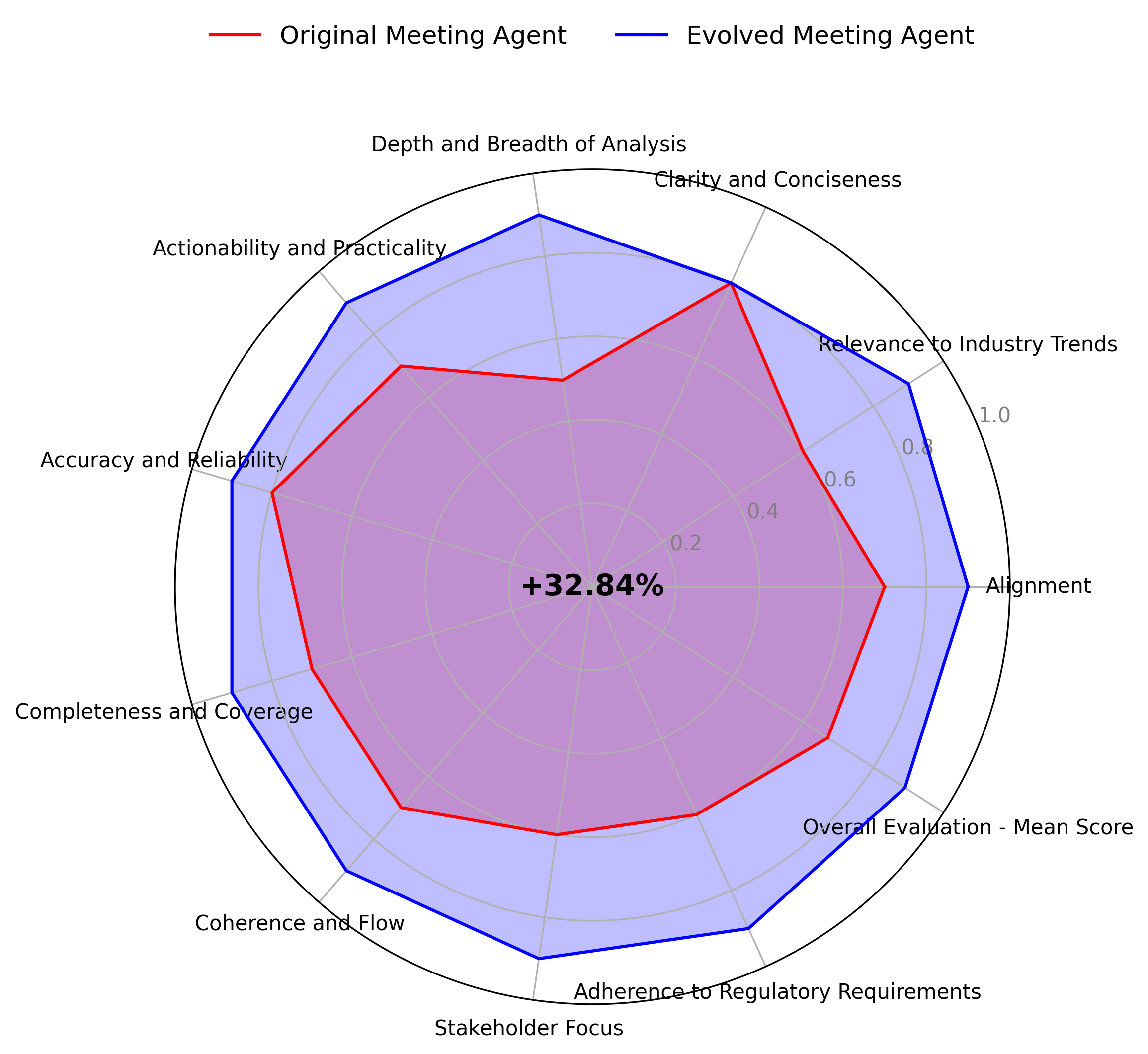}
\caption{Meeting Agent Refinement}
\end{figure}

\subsection{Lead Generation Agent}

The lead generation agent for the "AI for Personalized Learning" platform faced challenges regarding alignment with business objectives and data accuracy. These limitations hindered the system’s ability to generate valuable leads for the EdTech industry. New specialized roles were created to enhance the system, including Market Analyst and Business Development Specialist, to improve lead qualification processes and data integrity. The task structure was broadened to incorporate detailed analyses and actionable recommendations. The evolved agent system significantly improved evaluation criteria, including alignment with business objectives (91\%) and data accuracy (90\%). The enhancements underscore the importance of specialized roles and a structured approach in lead identification and qualification processes.

\begin{figure}[h]
\centering
\includegraphics[width=0.5\textwidth]{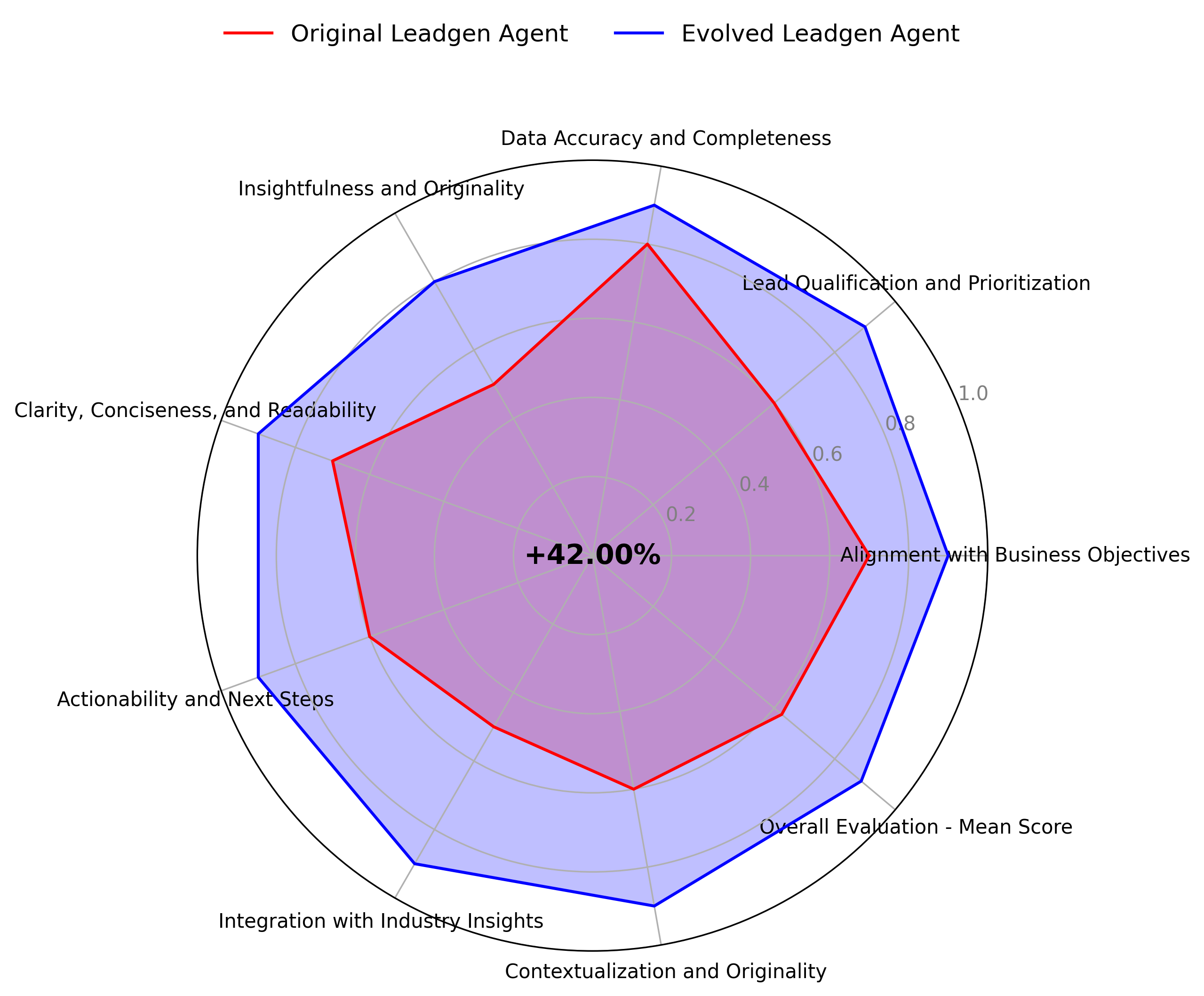}
\caption{Lead Generation Agent Refinement}
\end{figure}

\begin{figure*}[t]
\centering
\includegraphics[width=\textwidth]{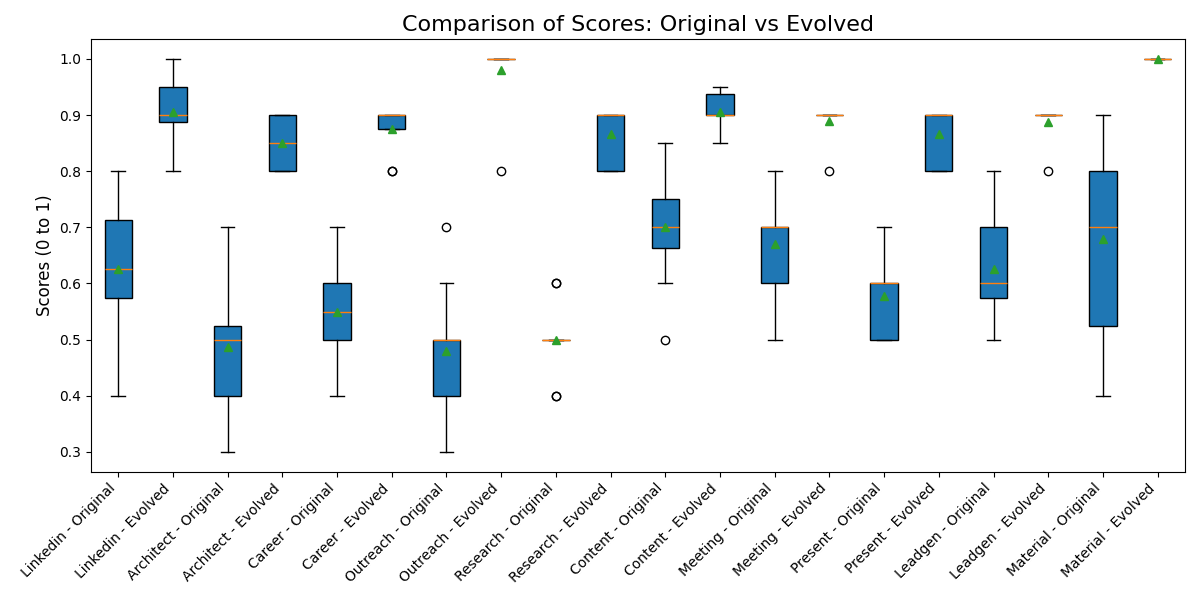}
\caption{Original vs Evolved System Comparisons across Multiple Case Studies: Each pair of bars represents the evaluation scores for original and evolved systems, highlighting significant alignment, clarity, relevance, and actionability improvements achieved by refining agents, tasks, and workflows. The evolved systems consistently demonstrate higher scores, indicating the effectiveness of introducing specialized roles and targeted modifications.}
\label{fig:comparison_boxplot}
\end{figure*}

\subsection{Evaluation Results}
Figure \ref{fig:comparison_boxplot} illustrates the comparative evaluation of original and evolved systems across various agentic applications, including customer support, medical imaging, supply chain management, and more. The box plots represent the distribution of evaluation scores on key criteria such as alignment, clarity, relevance, and actionability; with notable insights:

\begin{itemize}
    \item \textbf{Consistent Improvements:} The evolved systems achieve markedly higher scores across all case studies, with median values near or exceeding 0.9, demonstrating the benefits.
    \item \textbf{Variability Reduction:} The reduced spread in scores for evolved systems reflects more consistent and reliable outputs, attributable to the specialized agent roles and tasks.
    \item \textbf{Targeted Enhancements:} Systems such as the Outreach Agent, Market Research Agent, and Medical AI Architect showcase substantial improvements, highlighting the value of user-centric and data-driven approaches.
\end{itemize}

These findings underscore the transformative impact of continuous refinement in agent systems, emphasizing the importance of domain-specific roles, strategic modifications, and adaptability to meet the dynamic needs of industries and users.
\newpage
\section{Discussion}
The collective findings from case studies illustrate the transformative impact of targeted modifications and the introduction of specialized roles within agent systems. Each system's evolution resulted in substantial improvements across various evaluation criteria, including alignment, accuracy, relevance, clarity, and actionability. These enhancements not only addressed the initial challenges faced by the original systems but also positioned the evolved agent systems as valuable tools for their respective domains. The experiments conducted across these diverse case studies underscore the necessity for continuous refinement in agent systems to meet the evolving needs of industries and users. Introducing specialized roles and a user-centric design focus have proven instrumental in enhancing output quality and effectiveness. The insights gained from case studies will serve as a basis for future agent systems, emphasizing the importance of specialization and adaptability in achieving optimal results. A critical insight from \citet{sulc2024towards} is the importance of self-improving agents that adjust roles and interactions autonomously via feedback loops. Experiment results demonstrate the potential for dynamic adaptation and continuous enhancement, making it well-suited for environments with evolving objectives and conditions.

\section{Conclusion}
This paper presents a robust method for the autonomous refinement and optimization of Agentic AI solutions. The presented method continuously improves agent-based workflows by leveraging iterative feedback loops, hypothesis generation, and automated modifications, enhancing efficiency and effectiveness. The proposed method's autonomous nature minimizes human intervention, making it ideal for large-scale applications that require ongoing refinement. The method's scalability, flexibility, and ability to adapt to evolving objectives make it a powerful tool for optimizing complex AI agents. 

While this method demonstrates promising advancements in Agentic AI, several avenues for future exploration could further enhance its capabilities. Investigating the role of human-in-the-loop strategies can bridge fully autonomous operations and scenarios where nuanced human judgment may be beneficial, especially during the initial deployment or in environments with high uncertainty. This could lead to hybrid systems where human expertise augments autonomous agent decision-making, ensuring safety and reliability without compromising autonomy. Collaborations with industry partners will also help tailor the method to real-world needs, ensuring adaptability and impact.

\newpage
\section{Limitations}
The proposed framework for the autonomous refinement of Agentic AI systems has certain limitations that warrant consideration. Using LLMs for feedback, hypothesis formation, and evaluation may lead to inaccuracies, lack of explainability, and biases stemming from their training data. The framework's effectiveness relies on well-defined evaluation criteria. Poor or biased criteria can result in suboptimal refinements, as agents cannot independently identify missing dimensions. Minimal human involvement can be problematic in high-stakes or ambiguous tasks, where nuanced judgment and ethical considerations, such as privacy or unintended consequences, are crucial. Iterative processes like hypothesis generation and evaluation are computationally intensive, potentially limiting adoption in resource-constrained settings.

\bibliography{custom}

\appendix

\section{Report for a Refinement Iteration}
\label{sec:appendix}

\subsection{Initial Hypotheses and Justifications}

\subsubsection{Introducing Specialized Agents}

\textbf{Hypothesis:} Creating specialized agents for distinct tasks will enhance depth and specialization, resulting in more thorough and expert analyses.

\begin{itemize}
    \item \textbf{Market Identification Specialist Agent}
    \begin{itemize}
        \item \textbf{Role:} Market Identification Specialist
        \item \textbf{Goal:} Identify a wide range of potential markets using advanced search tools.
        \item \textbf{Tools:} SerperDevTool, WebsiteSearchTool
    \end{itemize}

    \item \textbf{Consumer Needs Analyst Agent}
    \begin{itemize}
        \item \textbf{Role:} Consumer Needs Analyst
        \item \textbf{Goal:} Analyze consumer needs using web scraping tools.
        \item \textbf{Tools:} ScrapeWebsiteTool
    \end{itemize}
\end{itemize}

\textbf{Rationale:} Specialized agents focusing exclusively on specific tasks will bring deeper knowledge and more targeted approaches, improving the precision and quality of market research.

\subsubsection{Tool Integration}

\textbf{Hypothesis:} Better utilization of available tools (SerperDevTool, WebsiteSearchTool, ScrapeWebsiteTool) by specialized agents will result in more comprehensive and data-driven analyses.

\textbf{Rationale:} Leveraging tools designed for specific purposes (search and scraping) will provide richer datasets and insights, producing a more robust market analysis.

\subsubsection{Redefining Existing Tasks}

\textbf{Hypothesis:} Redefining tasks to align with specialized roles will increase efficiency and clarity in the workflow, leading to better outcomes.

\begin{itemize}
    \item \textbf{Market Identification Task} (Assigned to Market Identification Specialist Agent)
    \begin{itemize}
        \item \textbf{Description:} Identify potential markets for the new product using search tools.
        \item \textbf{Expected Output:} A list of thoroughly researched potential markets.
    \end{itemize}

    \item \textbf{Consumer Needs Analysis Task} (Assigned to Consumer Needs Analyst Agent)
    \begin{itemize}
        \item \textbf{Description:} Analyze consumer needs in the identified markets using web scraping.
        \item \textbf{Expected Output:} A detailed report on consumer needs supported by data from web scraping.
    \end{itemize}
\end{itemize}

\textbf{Rationale:} Clear definition and reassignment of tasks will ensure each specialized agent can focus on their core activities, enhancing productivity and effectiveness.

\subsubsection{Creating a New Task for Comprehensive Validation}

\textbf{Hypothesis:} Adding a validation task to compile and confirm findings from specialized agents will ensure the final output is accurate and cohesive.

\begin{itemize}
    \item \textbf{Market Confirmation Task} (Dependent on Market Identification and Consumer Needs Analysis)
    \begin{itemize}
        \item \textbf{Description:} Validate and compile the final list of potential markets and their needs.
        \item \textbf{Expected Output:} A comprehensive and validated report on potential markets and consumer needs.
    \end{itemize}
\end{itemize}

\textbf{Rationale:} A final validation and compilation step will integrate insights from both specialized agents, ensuring the report is consistent and logically structured, thereby enhancing overall output quality.

\subsection{Revised Workflow}
\begin{enumerate}
    \item \textbf{Market Identification Specialist Agent} executes the Market Identification Task.
    \item \textbf{Consumer Needs Analyst Agent} performs the Consumer Needs Analysis Task.
    \item \textbf{Market Research Agent} consolidates the findings through the Market Confirmation Task and produces the final report.
\end{enumerate}

\subsection{Detailed Report Outlining the Rationale}

\subsubsection{Introducing Specialized Agents}
\begin{itemize}
    \item \textbf{Market Identification Specialist Agent}
    \begin{itemize}
        \item \textbf{Role:} Market Identification Specialist
        \item \textbf{Goal:} Identify a wide range of potential markets using advanced search tools.
        \item \textbf{Tools:} SerperDevTool, WebsiteSearchTool
        \item \textbf{Rationale:} This agent's specialization in identifying markets using advanced search tools will enhance the depth and precision of market identification, providing a stronger foundation for subsequent analysis.
    \end{itemize}

    \item \textbf{Consumer Needs Analyst Agent}
    \begin{itemize}
        \item \textbf{Role:} Consumer Needs Analyst
        \item \textbf{Goal:} Analyze consumer needs using web scraping tools.
        \item \textbf{Tools:} ScrapeWebsiteTool
        \item \textbf{Rationale:} By focusing exclusively on analyzing consumer needs using web scraping tools, this agent can generate more detailed and data-driven insights into consumer behavior and preferences.
    \end{itemize}
\end{itemize}

\subsubsection{Redefining Existing Tasks}
\begin{itemize}
    \item \textbf{Market Identification Task}
    \begin{itemize}
        \item \textbf{Description:} Identify potential markets for the new product using search tools.
        \item \textbf{Expected Output:} A list of thoroughly researched potential markets.
        \item \textbf{Agent:} Market Identification Specialist Agent
        \item \textbf{Tools:} SerperDevTool, WebsiteSearchTool
        \item \textbf{Rationale:} Assigning this task to the specialized agent ensures focused and comprehensive market identification using appropriate tools.
    \end{itemize}

    \item \textbf{Consumer Needs Analysis Task}
    \begin{itemize}
        \item \textbf{Description:} Analyze consumer needs in the identified markets using web scraping.
        \item \textbf{Expected Output:} A detailed report on consumer needs supported by data from web scraping.
        \item \textbf{Agent:} Consumer Needs Analyst Agent
        \item \textbf{Dependencies:} Market Identification Task
        \item \textbf{Tools:} ScrapeWebsiteTool
        \item \textbf{Rationale:} This specialized task leverages web scraping to provide deep consumer insights, ensuring that data accurately identifies and supports consumer needs.
    \end{itemize}
\end{itemize}

\subsubsection{Creating a New Task for Comprehensive Validation}
\begin{itemize}
    \item \textbf{Market Confirmation Task}
    \begin{itemize}
        \item \textbf{Description:} Validate and compile the final list of potential markets and their needs.
        \item \textbf{Expected Output:} A comprehensive and validated report on potential markets and consumer needs.
        \item \textbf{Agent:} Market Identification Specialist Agent
        \item \textbf{Dependencies:} Market Identification Task, Consumer Needs Analysis Task
        \item \textbf{Tools:} None
        \item \textbf{Rationale:} This final validation task ensures consistency and logical structuring of the integrated insights from both specialized agents, resulting in a more reliable and cohesive report.
    \end{itemize}
\end{itemize}

\subsection{Comprehensive Comparison Report}

\subsubsection{Evaluation of New Output vs. Best-Known Output}

\begin{itemize}
    \item \textbf{Potential Markets Identification:}
    \begin{itemize}
        \item \textbf{Best-Known Output:} Identified two markets (India B2C E-Commerce, Sustainable Steel).
        \item \textbf{New Output:} Identified seven markets (Health and Fitness, Sustainable Products, Smart Home, Elderly Care, Pet Care, Remote Work, Educational Tech.).
        \item \textbf{Evaluation:} The new output is more comprehensive, covering a broader range of markets.
    \end{itemize}

    \item \textbf{Consumer Needs Analysis:}
    \begin{itemize}
        \item \textbf{Best-Known Output:} Detailed for two markets.
        \item \textbf{New Output:} Detailed for seven markets, including market needs, growth drivers, and potential products.
        \item \textbf{Evaluation:} The new output provides a more comprehensive and structured analysis.
    \end{itemize}

    \item \textbf{Actionability:}
    \begin{itemize}
        \item \textbf{Best-Known Output:} Provides actionable insights for two markets.
        \item \textbf{New Output:} Provides actionable insights for seven markets.
        \item \textbf{Evaluation:} The new output offers more actionable insights due to its broader scope.
    \end{itemize}

    \item \textbf{Product Development Recommendations:}
    \begin{itemize}
        \item \textbf{Best-Known Output:} Clear recommendations for two markets.
        \item \textbf{New Output:} Clear recommendations for seven markets.
        \item \textbf{Evaluation:} The new output provides more comprehensive recommendations.
    \end{itemize}

    \item \textbf{Completeness and Coherence:}
    \begin{itemize}
        \item \textbf{Best-Known Output:} Completes essential steps for two markets.
        \item \textbf{New Output:} Completes essential steps for seven markets.
        \item \textbf{Evaluation:} The new output is more complete.
    \end{itemize}
\end{itemize}

\textbf{Conclusion:} The new output is superior to the best-known output as it provides:
\begin{enumerate}
    \item A broader and clearer identification of potential markets.
    \item A more comprehensive and structured consumer needs analysis.
    \item More actionable insights and recommendations for product development.
    \item Greater completeness and coherence in the market research process.
\end{enumerate}

Thus, the new variant (its code and output) has been saved as the best-known variant.

\end{document}